\documentclass[letterpaper, 10 pt, conference]{ieeeconf}  % Comment this line out if you need a4paper

\IEEEoverridecommandlockouts                              % This command is only needed if 
\usepackage[table, dvipsnames]{xcolor}
\usepackage[pdftex]{graphicx} % graphics package
\usepackage{graphicx}
\usepackage[lofdepth,lotdepth,caption=false]{subfig} % subplot package
\graphicspath{{figures/}} % declare the path where graphic files are
\DeclareGraphicsExtensions{.pdf,.jpg,.png} % grafic files extensions
\usepackage{amsmath} % math package for split
\usepackage{amsfonts} % math package for fonts
\usepackage{mathtools} % math package for tools like prescript
\usepackage{tikz} % figure package
\usetikzlibrary{shapes,arrows,fit,calc,positioning,automata,backgrounds,external}
\usepackage{color} % colors package
\usepackage{multirow} % multiple rows in table
\usepackage{hhline} % separator in table
\usepackage{enumerate} % enumerate package
\usepackage{epstopdf} % package to include eps. files
\usepackage{dsfont}
\usepackage{siunitx} % package for international system units
\usepackage{cite}   % for compact citations, i.e. [1, 2]
\usepackage{diagbox}    % for making slash cell in a table
\usepackage{comment}
\usepackage{algorithm}
\usepackage{algpseudocode}
\usepackage{epstopdf} % converting .eps to .pdf
\usepackage{soul} % for highlighting text

\definecolor{Bookcolor}{HTML}{00F9DE}
\definecolor{darkgreen}{rgb}{0.0, 0.5, 0.0}

\makeatletter
\let\NAT@parse\undefined
\makeatother

\usepackage[colorlinks=true,linkcolor=blue,urlcolor=blue,citecolor=blue,anchorcolor=blue]{hyperref} % hyperlink package (needs to be put at the end!)

\soulregister\ref7
\soulregister\eqref7
\soulregister\cite7
\soulregister\citep7
\soulregister\href7
\soulregister\equation7
\definecolor{highlightcolor}{rgb}{1,1,0.5}
\sethlcolor{highlightcolor}

\newtheorem{remark}{Remark}

\begin{document}

\title{PencilNet: Zero-Shot Sim-to-Real Transfer Learning \\for Robust Gate Perception in Autonomous Drone Racing}

% use a multiple column layout for up to three different affiliations
\author{Huy Xuan Pham, Andriy Sarabakha, Mykola Odnoshyvkin and Erdal Kayacan
% <-this % stops a space
\thanks{H. X. Pham and E. Kayacan are with the Artificial Intelligence in Robotics Laboratory~(Air Lab), Department of Electrical and Computer Engineering, Aarhus University, 8000 Aarhus C, Denmark {\tt\small \{huy.pham, erdal\} at ece.au.dk}}%
\thanks{A. Sarabakha is with the School of Electrical and Electronic Engineering~(EEE), Nanyang Technological University (NTU), Singapore, 639798. {\tt\small andriy.sarabakha at ntu.edu.sg}}%
\thanks{M. Odnoshyvkin and A. Sarabakha are with the Munich School of Robotics and Machine Intelligence, Technical University of Munich~(TUM), D-80797 Munich, Germany. {\tt\small mykola.odnoshyvkin at tum.de}}%
}


%%\author{\IEEEauthorblockN{Michael Shell\IEEEauthorrefmark{1},
%%Homer Simpson\IEEEauthorrefmark{2},
%%James Kirk\IEEEauthorrefmark{3}, 
%%Montgomery Scott\IEEEauthorrefmark{3} and
%%Eldon Tyrell\IEEEauthorrefmark{4}}
%%\IEEEauthorblockA{\IEEEauthorrefmark{1}School of Electrical and Computer Engineering\\
%%Georgia Institute of Technology,
%%Atlanta, Georgia 30332--0250\\ Email: see http://www.michaelshell.org/contact.html}
%%\IEEEauthorblockA{\IEEEauthorrefmark{2}Twentieth Century Fox, Springfield, USA\\
%%Email: homer@thesimpsons.com}
%%\IEEEauthorblockA{\IEEEauthorrefmark{3}Starfleet Academy, San Francisco, California 96678-2391\\
%%Telephone: (800) 555--1212, Fax: (888) 555--1212}
%%\IEEEauthorblockA{\IEEEauthorrefmark{4}Tyrell Inc., 123 Replicant Street, Los Angeles, California 90210--4321}}


\maketitle

% for page numbering. REMOVE IT FOR THE FINAL SUBMISSION
%\thispagestyle{plain}
%\pagestyle{plain}

% insert page header and footer here for IEEE PDF Compliant
%\thispagestyle{fancy}
%\fancyhead{}
%\lhead{}
%\lfoot{}
%\cfoot{}
%\rfoot{}
%\renewcommand{\headrulewidth}{0pt}
%\renewcommand{\footrulewidth}{0pt}

\begin{abstract}

In autonomous and mobile robotics, one of the main challenges is the robust on-the-fly perception of the environment, which is often unknown and dynamic, like in autonomous drone racing. 
In this work, we propose a novel deep neural network-based perception method for racing gate detection -- PencilNet\footnote{Source code, trained models, and collected datasets are available at \url{https://github.com/open-airlab/pencilnet}.} -- which relies on a lightweight neural network backbone on top of a pencil filter. This approach unifies predictions of the gates' 2D position, distance, and orientation in a single pose tuple. We show that our method is effective for zero-shot sim-to-real transfer learning that does not need any real-world training samples. Moreover, our framework is highly robust to illumination changes commonly seen under rapid flight compared to state-of-art methods. A thorough set of experiments demonstrates the effectiveness of this approach in multiple challenging scenarios, where the drone completes various tracks under different lighting conditions.

\end{abstract}
\IEEEpeerreviewmaketitle

%\bstctlcite{IEEEexample:BSTcontrol}

% introduction and related work
\section{Introduction}

Vision-based agile navigation for robotics is an emerging research field that is gaining traction in recent years~\cite{andersen2022event, camci2019learning, Blosch2010ICRA, 9143655, 9196964}. As technical difficulties hinder the extension of batteries' capacity, there is a strong interest in increasing the speed of robots to expand both operating range and capabilities of the systems~\cite{Wang2021IROS}. This is even more crucial for aerial systems, like multicopters~\cite{Patel2019JIRS}, that are capable of reaching high speeds in a short time due to their agile nature~\cite{Patel2019REDUAS}. Significant signs of progress for faster quadrotor flights have been made through addressing drone racing~\cite{Moon2019ISR}, previously thought of as merely an entertainment application. Although, autonomous drone racing is an excellent playground for the developed drone technologies which can be transferred to other domains~\cite{Pham2022Book}, such as search and rescue. In autonomous drone racing, a drone has to fly autonomously and safely through a track that consists of multiple racing gates~\cite{DeWagter2021NMI}.

Early works~\cite{jung2018direct, li2020autonomous, rojas2017metric} normally use hand-crafted gate perception on top of traditional control and planning frameworks to achieve safe gate passing. However, they are less robust to real-world conditions~\cite{de2021artificial}. Subsequent work tends to exploit deep neural networks (DNNs) that are more effective to cope with perception uncertainties~\cite{Morales2020IJCNN}. A variant ResNet-8 DNN with three residual blocks is proposed in~\cite{kaufmann2019beauty} explicitly estimates a gate distribution. The method in~\cite{foehn2020alphapilot} utilizes a variant of \mbox{U-Net}~\cite{ronneberger2015u} that provides gate segmentation and demonstrates robust performance on top of a well-engineered system. The approach in~\cite{de2021artificial} also segments the gates with U-Net but associates the gate corners differently by combining corner pixel searching and re-projection error rejection. Recently, end-to-end methods~\cite{loquercio2018dronet, song2021autonomous, pfeiffer2022visual} using a single DNN that outputs tracking targets for controllers are considered to reduce the latency introduced by traditional modulation of perception, planning, and control sub-problems. In both lines of work, the quality of a gate perception, either through explicit mapping of gates or implicit feature-based understanding, dictates the performance of the system overall.

\begin{figure}[!t] \label{fig:abstract}
\centering
    {\includegraphics[width=0.9\linewidth]{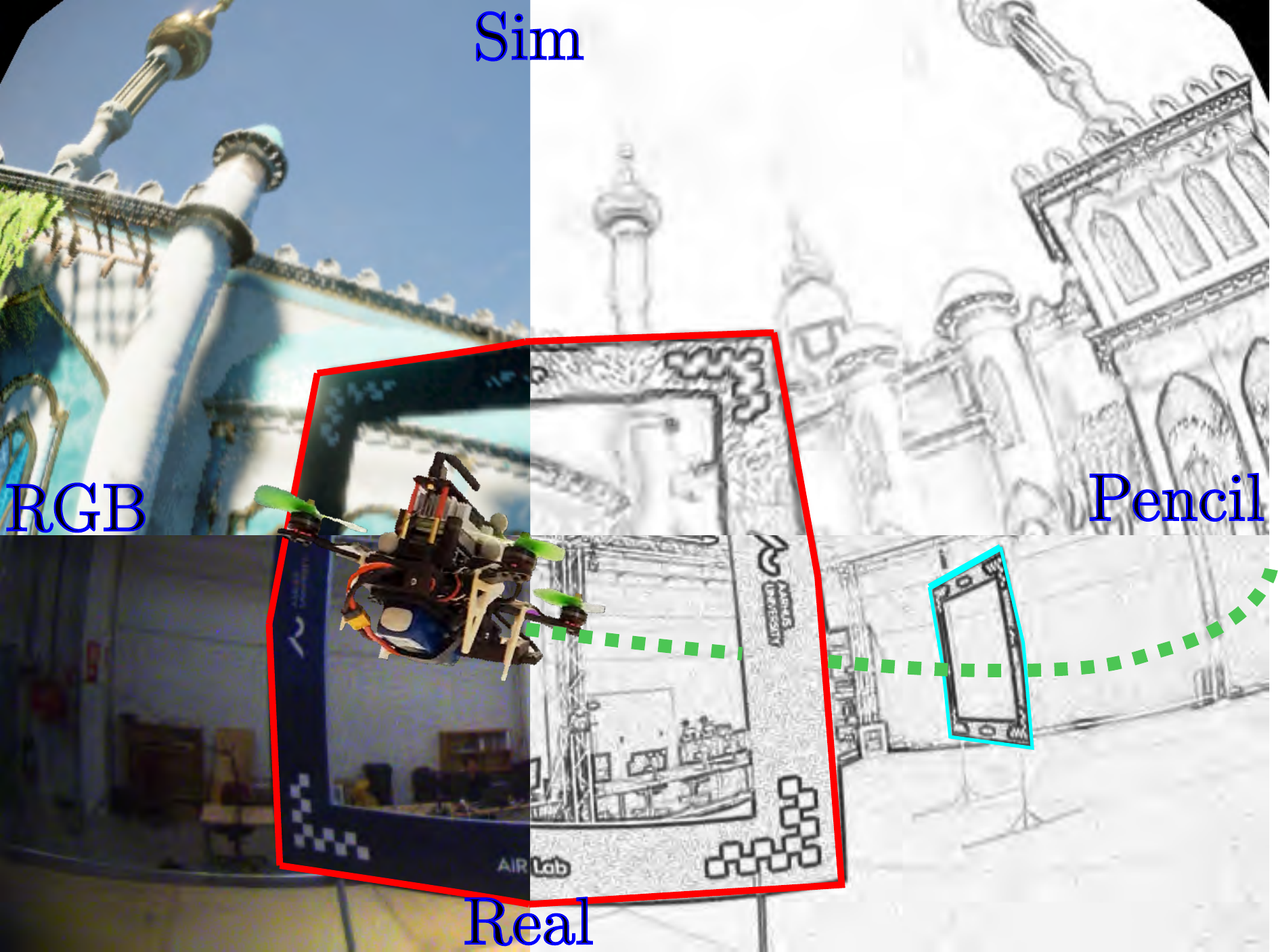}} 
    \caption{Composed image of simulated (top left) and real (bottom left) gates and converted counterparts (right-hand side) with pencil filter. The perception network is trained solely on simulation data and can accurately perceive racing gates in real-world environments through a fish-eye camera.}
    \vspace{-0.2cm}
\end{figure}

Another point to take into consideration is the ease of training the perception network for drone racing. Unlike many robotics applications, the cost to collect labelled high-speed real-world data for drone racing is often prohibitive which limits the use of supervised methods. Therefore, learning through simulation data is understandably much desired~\cite{zakharov2019dpod}. For this purpose,~\cite{loquercio2018dronet} uses domain randomization for sim-to-real transfer learning with robust performance. A reinforcement learning method is proposed in~\cite{song2021autonomous} to learn in simulation a policy for drones that fly at high speeds but reported high tracking errors in practice due to the abstraction from input data to real physical drones is far more complex in reality than in simulation. Unfortunately, none of these methods has been extensively tested under various real-world conditions, such as illumination alterations, which are strongly relevant for drone racing and widely known to negatively affect the performance of perception networks.

In this work, we propose a hybrid DNN-based approach that demonstrates an effective zero-shot sim-to-real transfer learning capability for gate perception in autonomous drone racing navigation. By relying on a smarter input representation using a low-cost morphological operation, called pencil filtering, we can bridge the sim-to-real gap so that our perception network is trained solely in simulation data and works robustly in real-world racing tracks, even under rapid motions and drastic lighting changes. The proposed framework attains a high-speed inference due to the use of our previously proposed efficient backbone network~\cite{gatenethuypham2021} with a modification to allow it to work with the new input representation. In our experiments, we show that the performance of the network significantly outperforms the original backbone network~\cite{gatenethuypham2021} and other state-of-the-art baseline methods in terms of accuracy and robustness. 

The rest of this work is organized as follows. The proposed method for gate perception is detailed in Section~\ref{sec:methodology}. Section~\ref{sec:evaluation} presents experiments to evaluate the efficiency of the approach against the state-of-the-art baselines. A real-world validation study is presented in Section \ref{sec:experiment}. Finally, Section~\ref{sec:conclusion} concludes this work.

\section{Proposed Method}
\label{sec:methodology}

\subsection{Pencil Filter}

Drastic illumination changes and blurriness caused by rapid flights have been extensively documented in previous work as the main factors behind the performance degradation of visual-based navigation frameworks. In this work, we tackle these challenges by employing an edge-enhancing technique known as the pencil filter~\cite{rambach2018learning}, but unlike~\cite{rambach2018learning}, we study its performance under various light intensities and motion blur. The pencil filter applied to an image emphasizes geometrical features, such as edges, corners, and outlines but preserves the overall image structure. This reduces the real-to-sim gap since images from simulation lack accuracy in objects' sharpness and illumination compared to real images, allowing learning on synthetic data and applying in the real-world environment. In addition, the pencil filter is an illumination agnostic method that keeps the light intensity constant in the images. To extract geometrical features pencil filter first converts an RGB image to a gray-scale format. Then, it applies dilation by convolving the gray-scale image with an ellipse kernel and thresholding to obtain local maxima for each pixel in its neighborhood. This morphological operation generally exaggerates the bright regions of the image.  To avoid information loss in the dark regions, the difference in pixel values of the gray-scale image and the dilated one is calculated by normalizing their respective values. The resulting image retains sharp and bolder edges, accentuating object boundaries. Thus, sim-to-real transfer performance is strongly enhanced by converting training RGB images into pencil images. The resulting pencil image is shown on the right-hand side of Fig.~\ref{fig:abstract}; while the pseudo-code of the pencil filter is provided in Algorithm~\ref{alg:pencil_filter}.

\algnewcommand\algorithmicforeach{\textbf{for each}}
\algdef{S}[FOR]{ForEach}[1]{\algorithmicforeach\ #1\ \algorithmicdo}
\begin{algorithm}[!t]
    \caption{Pencil filter.}
    \label{alg:pencil_filter}
    \begin{algorithmic}[1]
        \State{\textbf{Input:} RGB image $\mathbf{I}$}
        \State{\textbf{Output:} pencil image $\mathbf{P}$}
        \Function{pencil\_filter}{$\mathbf{I}$}
            \State{$\mathbf{G} \gets \text{grayscale}(\mathbf{I})$}
            \State{$\mathbf{P} \gets \text{dilate}(\mathbf{G}, ellipse)$}
            \ForEach{$y \in [1, \text{rows}(\mathbf{P})]$}
                \ForEach{$x \in [1, \text{columns}(\mathbf{P})]$}
                    \If{$\mathbf{P}(y,x) = 0$}
                        \State{$\mathbf{P}(y,x) \gets 255$}
                    \Else
                        \State{$\mathbf{P}(y,x) \gets  \text{int} \left( 255 \dfrac{\mathbf{G}(y,x)}{\mathbf{P}(y,x)} \right)$}
                    \EndIf
                \EndFor
            \EndFor
            \State{\Return $\mathbf{P}$}
        \EndFunction
    \end{algorithmic}
\end{algorithm}

\subsection{Training Data Generation}
\label{sub:sim_data_generation}

\begin{figure}[!b]
    \centering
    % \resizebox{0.49\textwidth}{!}{
    % \input{large_elements/fig_simulation_pipeline}
    % }
    \includegraphics[width=0.47\textwidth]{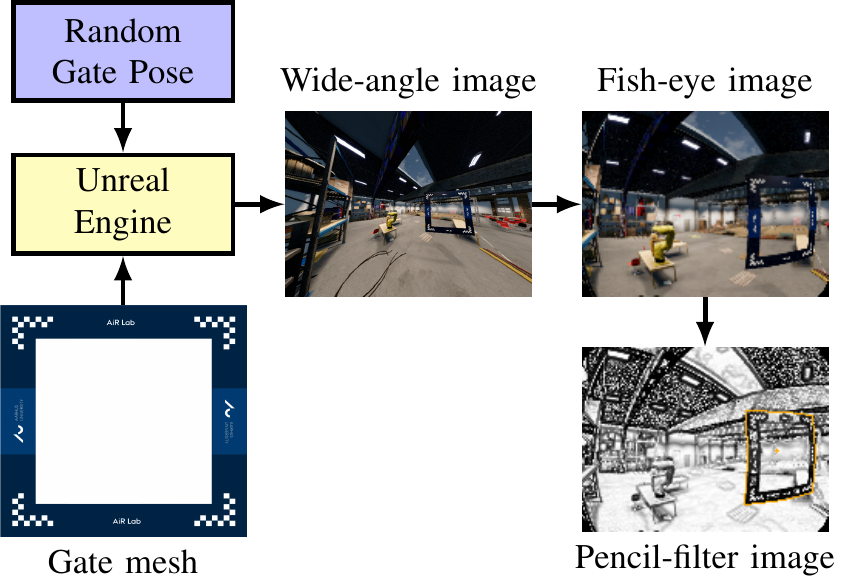}
    \caption{Synthetic images generation pipeline: first, gates are randomly spawned in the simulation environment; then, wide-angle images are distorted with a fish-eye model; finally, the pencil filter is applied.}
    \label{fig:simulation_generation}
\end{figure}

The key feature to reduce the gap between simulation and real-world appearance is the use of photo-realistic images. Graphical gaming engines allow simulating intricate ambient effects, mimicking the visual appearance of real environments and the surrounding noise and perturbations. In this work, photo-realistic images are generated by Unreal Engine containing racing scenes with vertical gates in multiple environments with different conditions. Similar to previous drone racing studies in~\cite{foehn2020alphapilot,gatenethuypham2021}, we use square-shaped gates with checkerboard-like patterns around each corner. To generate a large number of distinct images, a wide-angle RGB camera and the gates are spawned randomly in the environment. They have to be distorted and interpolated to obtain simulated images with the same camera parameters as those from a real camera with a fish-eye lens. Finally, the pencil filter is applied to the fish-eye images. The pipeline for generating training images is depicted in Fig.~\ref{fig:simulation_generation}.

\begin{figure}[!b]
    \centering
    \subfloat[Synthetic images after applying fish-eye distortion.]{\includegraphics[width=0.49\textwidth]{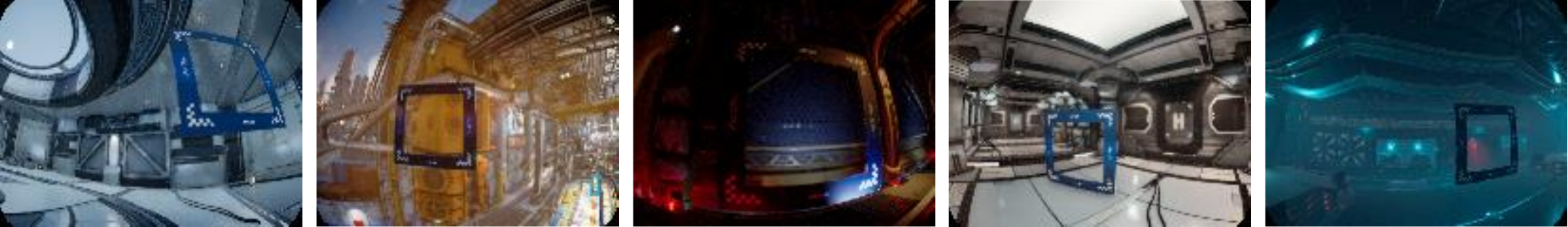} \label{fig:sim_fisheye}} \hfill
    \subfloat[Synthetic images after applying pencil filter.]{\includegraphics[width=0.49\textwidth]{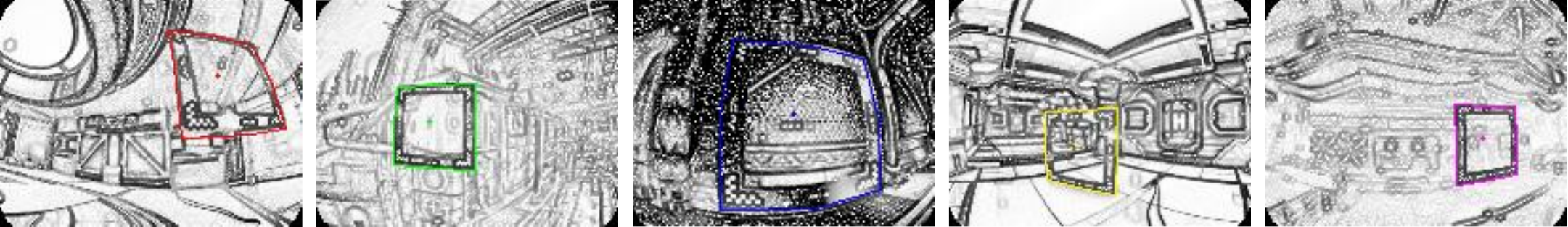} \label{fig:sim_pencil}}
    \caption{Samples of the simulation data before (\ref{fig:sim_fisheye}) and after (\ref{fig:sim_pencil}) the pencil filter are applied. The images are generated from different domains with varying angles and illumination settings. Note that the training images do not contain bounding boxes of gates, which are illustrated here merely for better visualization.}
    \label{fig:simulation_data}
\end{figure}

\begin{figure}[!t]
\centering
\subfloat[Synthetic training dataset.]{\includegraphics[width=0.47\textwidth]{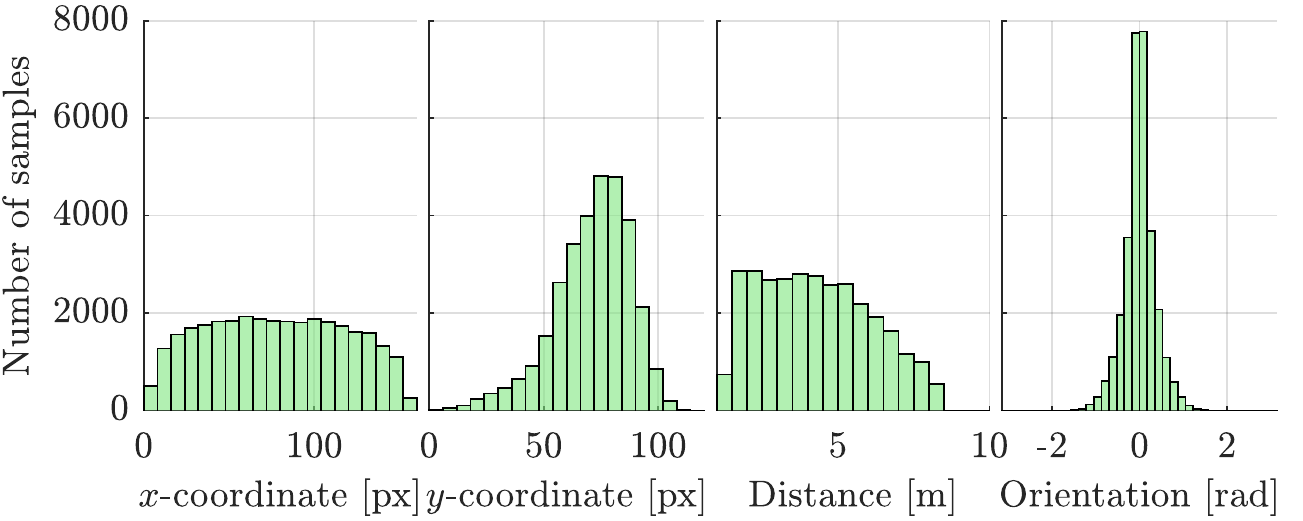} \label{fig:histogram_train}} \hfill
\subfloat[Synthetic testing dataset.]{\includegraphics[width=0.47\textwidth]{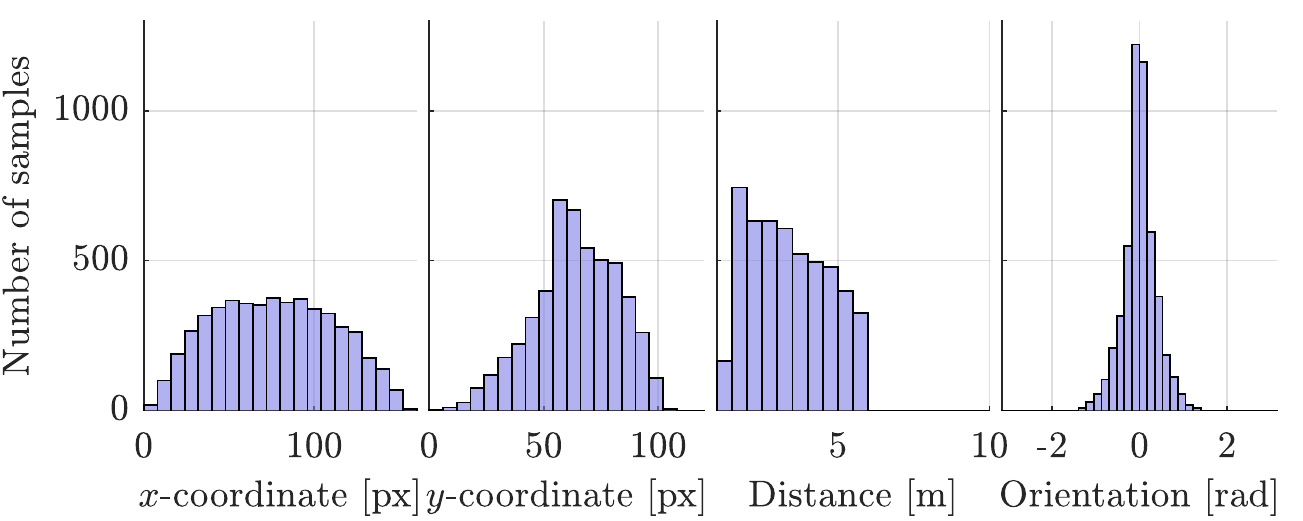} \label{fig:histogram_test}} \hfill
\subfloat[Real-world testing dataset.]{\includegraphics[width=0.47\textwidth]{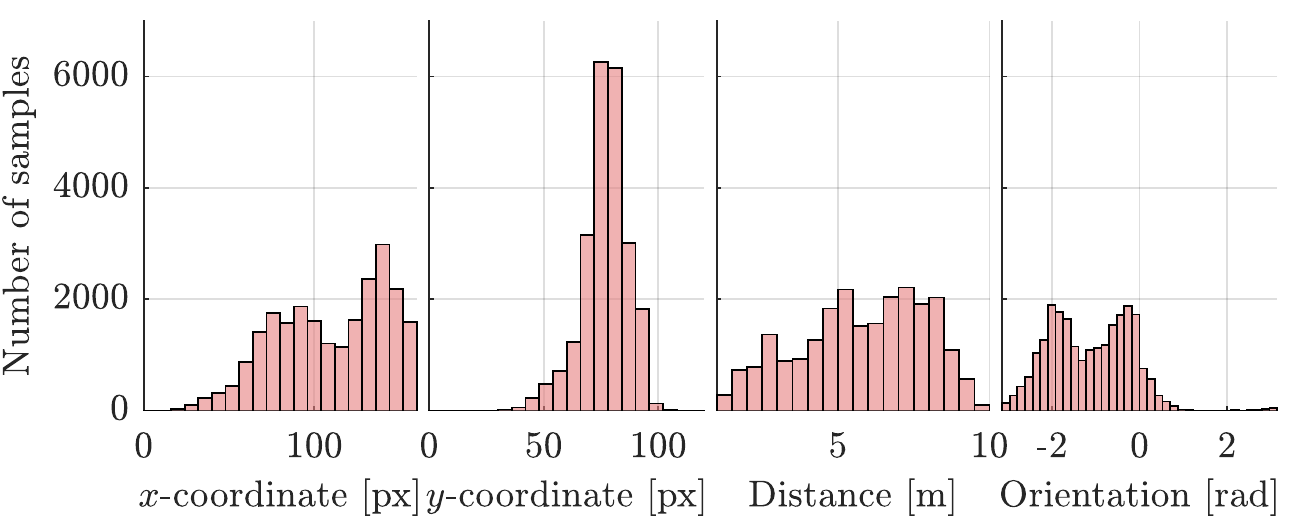} \label{fig:histogram_real}}
\caption{Distributions of $x$ and $y$-coordinates, distance and orientation of the gate on images in training and testing datasets.}
\label{fig:histogram}
\end{figure}

Using ground-truth information from the simulator, the RGB fish-eye images are labeled with five variables as a tuple $\left\{ x, y, d, \theta, c \right\}$. $x$ and $y$ are normalized pixel differences between the top-left corner of the grid and the gate's center, $d$ is the relative distance to the gate (in meters), and $\theta$ is the relative orientation of the gate (in radians), calculated as the difference between gate's yaw (heading) and camera's yaw projected onto the horizontal plane. $c$ is the confidence value of a corresponding grid, which has a value of $1$, if a gate center locates inside, and $0$, otherwise. 
The training dataset includes more than $30$K samples drawn from 13 different environments with 18 distinct backgrounds and various lighting conditions. Each data sample consists of annotations of the gates and a $160\times120$ RGB image (illustrated in Fig.~\ref{fig:sim_fisheye}). We do not apply any augmentation to the gates and images. In the training phase, the RGB images are converted into corresponding pencil-filtered images (illustrated in Fig.~\ref{fig:sim_pencil}), which retain the same annotations. The advantage of using simulation for data collection is that the data distributions are fully controlled, as shown in Fig.~\ref{fig:histogram_train}. Note that the training data only contain labels for the front side of a gate, as the back side does not provide relevant information.

\subsection{Backbone Detection Network}

\begin{figure*}[!th]
    \centering
    \includegraphics[width=0.99\textwidth]{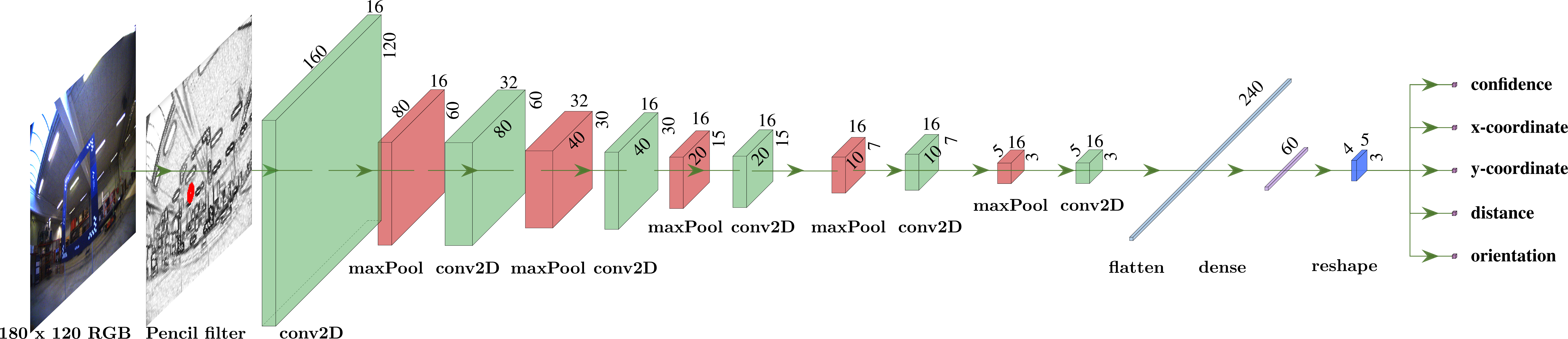}
    \caption{Architecture of the proposed method: PencilNet. The pencil filter is applied to RGB images before feeding them to the first conv2D layers. The backbone network is similar to~\cite{gatenethuypham2021}, with the modification in reducing the number of channels for each filter to work with one-channel input. Each of the first five conv2D layers is followed by a max-pooling with a $2\times2$ kernel. The tensor in the last conv2D layer is flattened to increase connectivity with the following dense layer. The output tensor is reshaped as a grid of $4\times3$ cells, where each cell contains five prediction values for pixel offsets to the gate's center, relative distance and orientation of the gate, and a confidence value.}
    \label{fig:gatenet}
\end{figure*}

The detection network takes a $160\times120$ image converted by the pencil filter as input before normalization. For feature detection, PencilNet, illustrated in Fig. \ref{fig:gatenet}, utilizes as the backbone a convolutional neural network~(CNN), from our previous work~\cite{gatenethuypham2021}, comprising of six 2D convolutional (conv2D) layers with a small number of filters in each layer and one fully-connected (dense) layer to regress detection outputs. The tensors of PencilNet are modified to process one-channel images leading to a smaller number of parameters. Each conv2D layer is followed by a batch normalization and a rectified linear unit~(ReLU) activation. Besides, the first five layers use max-pooling with $2\times2$ kernels. In the last conv2D layer, a small $3\times5\times16$ tensor is flattened so that each hidden neuron in the following dense layer can be connected with the extracted features.
As in \cite{gatenethuypham2021}, the network output is reshaped to $R \times C \times F$, where each grid cell from $R = 4$ rows and $C = 3$ columns contains regressed $F = 5$ values of $\{x, y, d, \theta, c\}$. 

\subsection{Training}
Similarly to~\cite{gatenethuypham2021}, PencilNet uses four losses $\mathcal{L}_{xy}$, $\mathcal{L}_{d}$, $\mathcal{L}_{\theta}$ and $\mathcal{L}_{c}$, which are center coordinate, distance, orientation and confidence deviations from true values, respectively:
\begin{equation}
    \centering
    \begin{cases}
        \mathcal{L}_{xy} &= { {
            \sum_{i = 1}^{R}
                \sum_{j = 1}^{C}      
                  {\mathds{1}}_{ij}^{\text{obj}}
            \left[
            \left(
                x_{ij} - \hat{x}_{ij}
            \right)^2 +
            \left(
                y_{ij} - \hat{y}_{ij}
            \right)^2
            \right],
        }}\\
        \mathcal{L}_{d} &= { {
            \sum_{i = 1}^{R}
                \sum_{j = 1}^{C}      
            {\mathds{1}}_{ij}^{\text{obj}} 
            \left(
                d_{ij} - \hat{d}_{ij}
            \right)^2,
        }}\\
        \mathcal{L}_{\theta} &= { {
            \sum_{i = 1}^{R}
                \sum_{j = 1}^{C}      
            {\mathds{1}}_{ij}^{\text{obj}}
            \left(
                \theta_{ij} - \hat{\theta}_{ij}
            \right)^2,
        }}\\
        \mathcal{L}_{c} &= { {
            \sum_{i = 1}^{R}
                \sum_{j = 1}^{C}      
                  \left[
                  {\mathds{1}}_{ij}^{\text{obj}}+\alpha (1-{\mathds{1}}_{ij}^{\text{obj}})\right]
            \left(
                c_{ij} - \hat{c}_{ij}
            \right)^2.
        }}\\
    \end{cases}
    \label{eq:loss_function}
\end{equation}
The losses are calculated for each grid cell of the output layer with row and column indices $i\in[1,R]$ and $j\in[1,C]$. Terms with the hat operator $(\hat{\star})$ denote predicted values. %Again, $(x, y)$ are pixel offsets from the gate center to the top left corner of each grid cell along the $x$ and $y$ axes, $d$ is the relative distance from the camera to a gate center, $\theta$ is the relative orientation of the gate, and $c$ is the confidence value. 
The term ${\mathds{1}}_{ij}^{\text{obj}}$ is a binary variable to penalize the network only when a gate center locates in a particular grid. The confidence values for grid cells where a gate center does not present is minimized with a weight $\alpha$ so that in run time we can threshold the low confidence predictions, thus reducing false positives. Finally, the loss function is a weighted term:
 \begin{equation}
\centering
 \mathcal{L} = \lambda_{xy} \mathcal{L}_{xy} + \lambda_{d} \mathcal{L}_{d} + \lambda_{\theta} \mathcal{L}_{\theta} + \lambda_{c}\mathcal{L}_{c},
 \end{equation}
 where $\lambda_{xy}$, $\lambda_{d}$, $\lambda_{\theta}$, $\lambda_{c}$ are weights reflecting importance for each of the losses.

\section{Gate Detection Evaluation}
\label{sec:evaluation}

\subsection{Baseline Methods}

In the first part of our study, we evaluate PencilNet performance for the task of racing gate perception with a few notable milestone methods that include DroNet variants (used in~\cite{kaufmann2019beauty} and~\cite{loquercio2019deep}), and ADRNet variants~\cite{jung2018perception}. We also include the performance of our previously published GateNet~\cite{gatenethuypham2021}, to evaluate the contribution of the pencil filter. 

\remark{The above baselines are chosen because of their similarities to our method. Some other state-of-the-art baselines such as~\cite{foehn2020alphapilot} and~\cite{de2021artificial} are not included in this study due to the differences in training data and target outputs.}

\subsubsection{{DroNet variants \cite{kaufmann2019beauty, loquercio2018dronet}}} A variant of \mbox{ResNet-8 DNN} with three residual blocks is proposed in~\cite{loquercio2018dronet} to originally learn a steering policy for a quadrotor to avoid obstacles. However, due to its efficient structure, many subsequent papers utilize it as a backbone feature extractor with different output layers. Kaufmann et al.~\cite{kaufmann2019beauty} replace the DroNet's original output layers with multi-layer perceptron to estimate the mean and variance of the next gate’s pose expressed in spherical coordinates and use the predicted pose for a model predictive control framework to compute a trajectory for the drone. Loquercio et al.~\cite{loquercio2018dronet} train DroNet with simulation data using domain randomization and also change the output layers to produce a velocity vector that drives the robot through the gate. Pfeiffer et al.~\cite{pfeiffer2022visual} combine DroNet with an attention map as an end-to-end network to capture visual-spatial information allowing the system to track high-speed trajectories. In this work, we compare PencilNet with two variants of DroNet: DroNet-1.0 with the full set of parameters as in~\cite{kaufmann2019beauty}, and DroNet-0.5 with only half of the filters as in~\cite{loquercio2018dronet}. The original output layers are also modified to produce similar outputs as our method for compatibility reasons.

\subsubsection{{ADRNet variant~\cite{jung2018perception}}}
A single-shot object detector based on AlexNet~\cite{krizhevsky2012imagenet} is proposed in~\cite{jung2018perception} to detect the bounding box of a gate in an image frame, which is used to calculate offset velocities needed to guide a drone past the gate. Since the original ADRNet only detects the gate center and lacks distance and orientation estimation, this work also replaces its output layer with our output layer to predict additional attributes (named ADRNet-mod).

\subsubsection{{GateNet variants~\cite{gatenethuypham2021}}}

\begin{figure}[!b]
\centering
\subfloat[]{\includegraphics[width=0.097\textwidth]{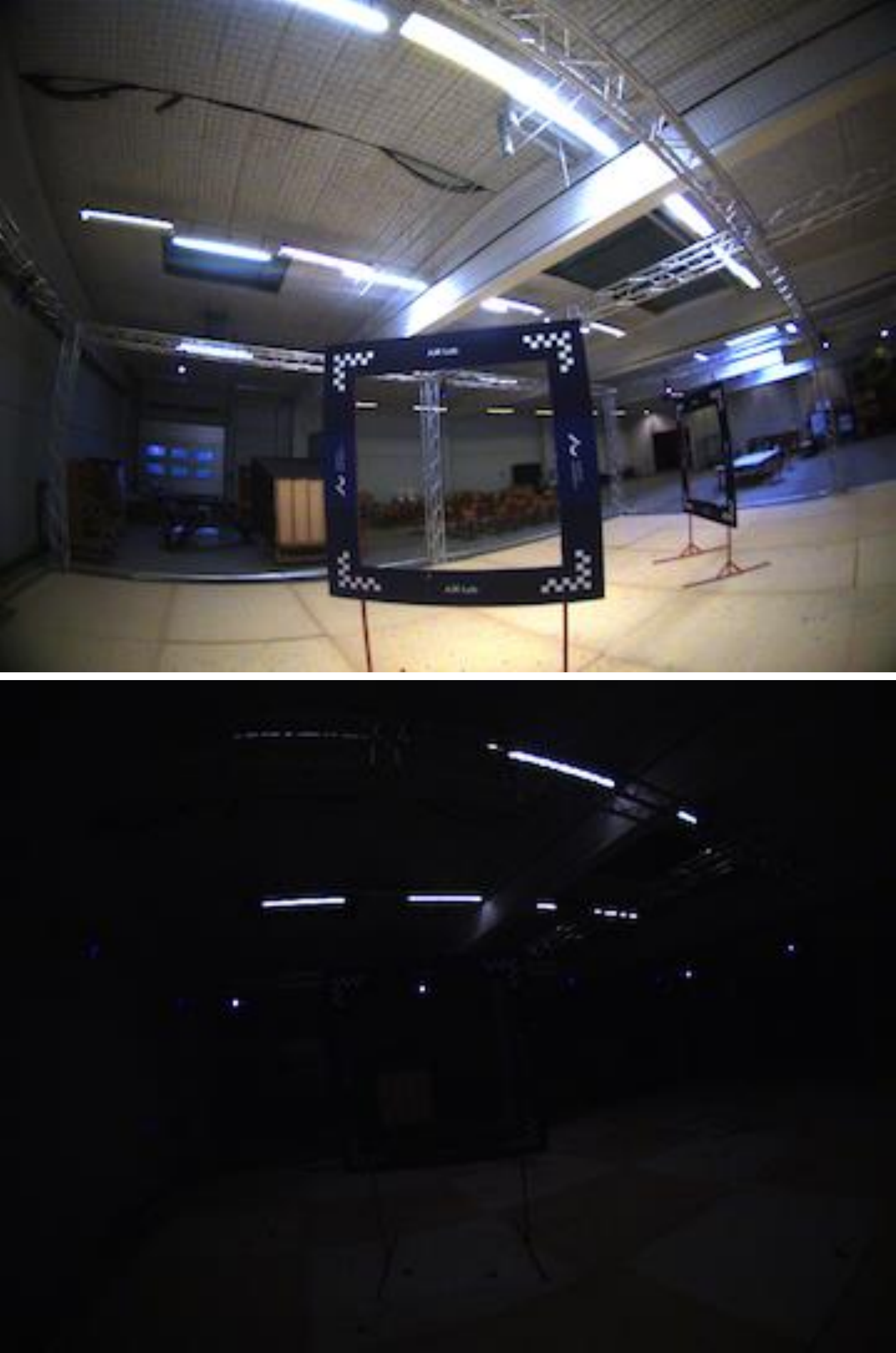}\label{fig:filters_0}}\hfill%
\subfloat[]{\includegraphics[width=0.097\textwidth]{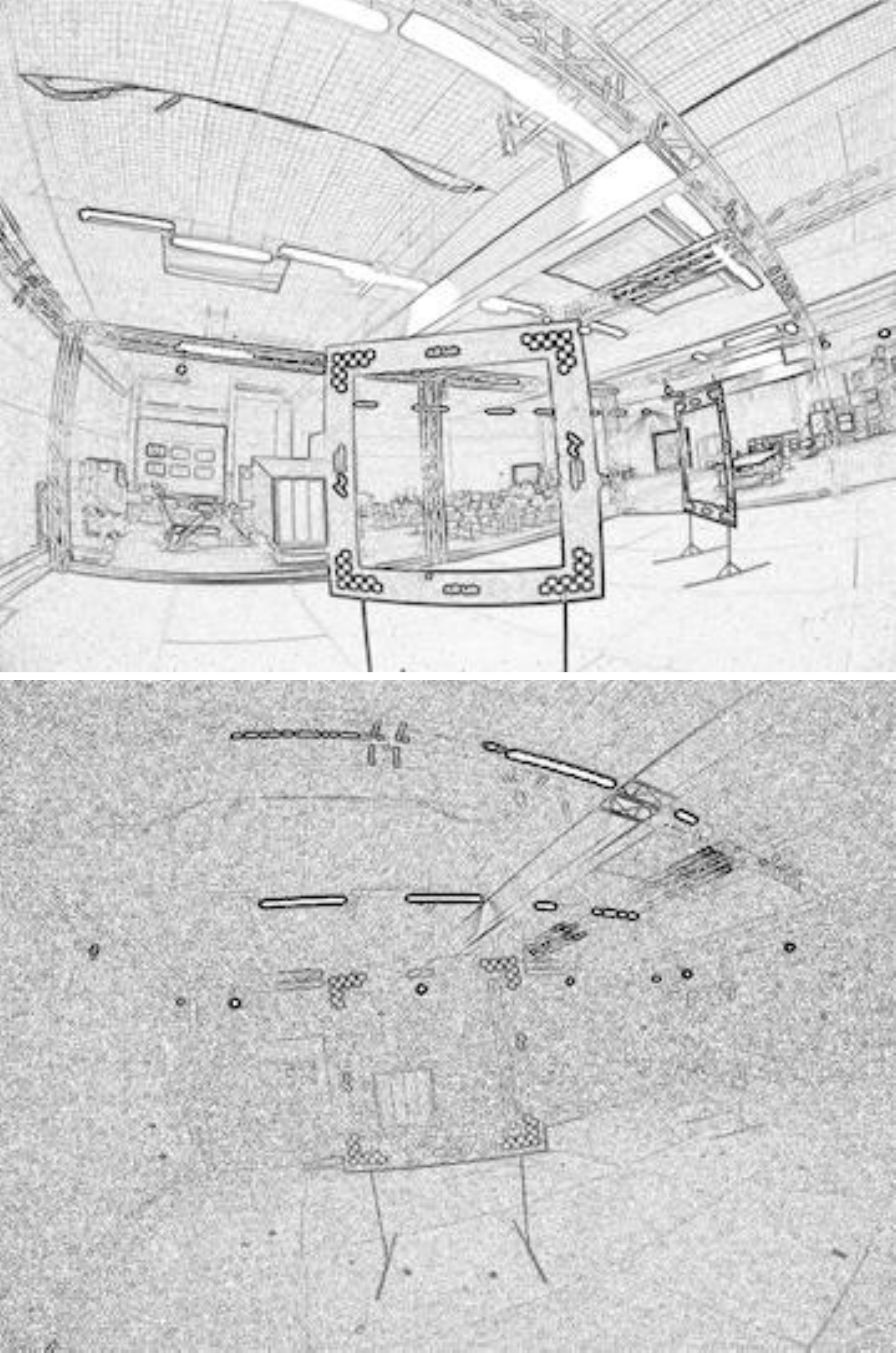}\label{fig:filters_1}}\hfill%
\subfloat[]{\includegraphics[width=0.097\textwidth]{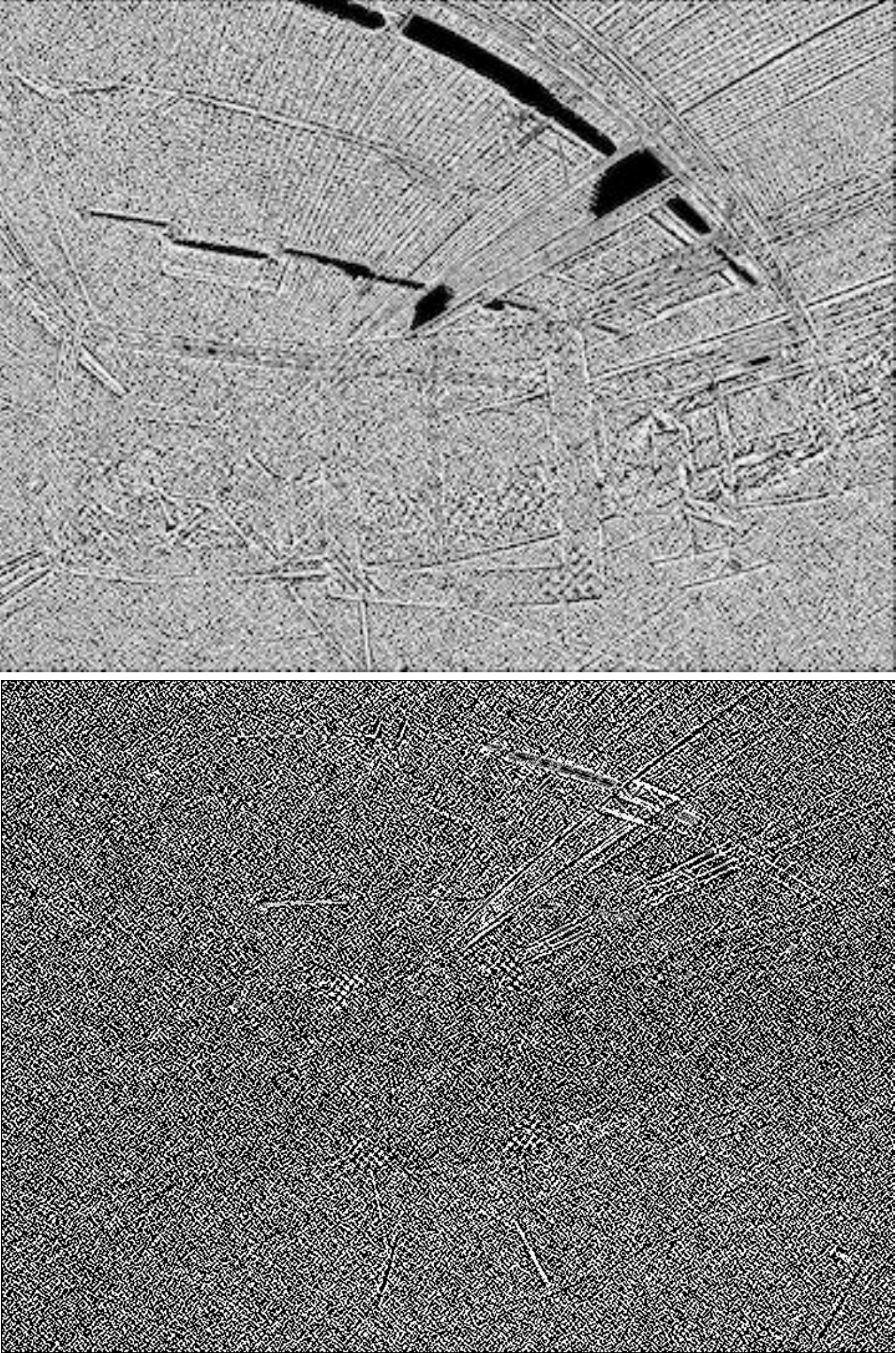}\label{fig:filters_2}}\hfill%
\subfloat[]{\includegraphics[width=0.097\textwidth]{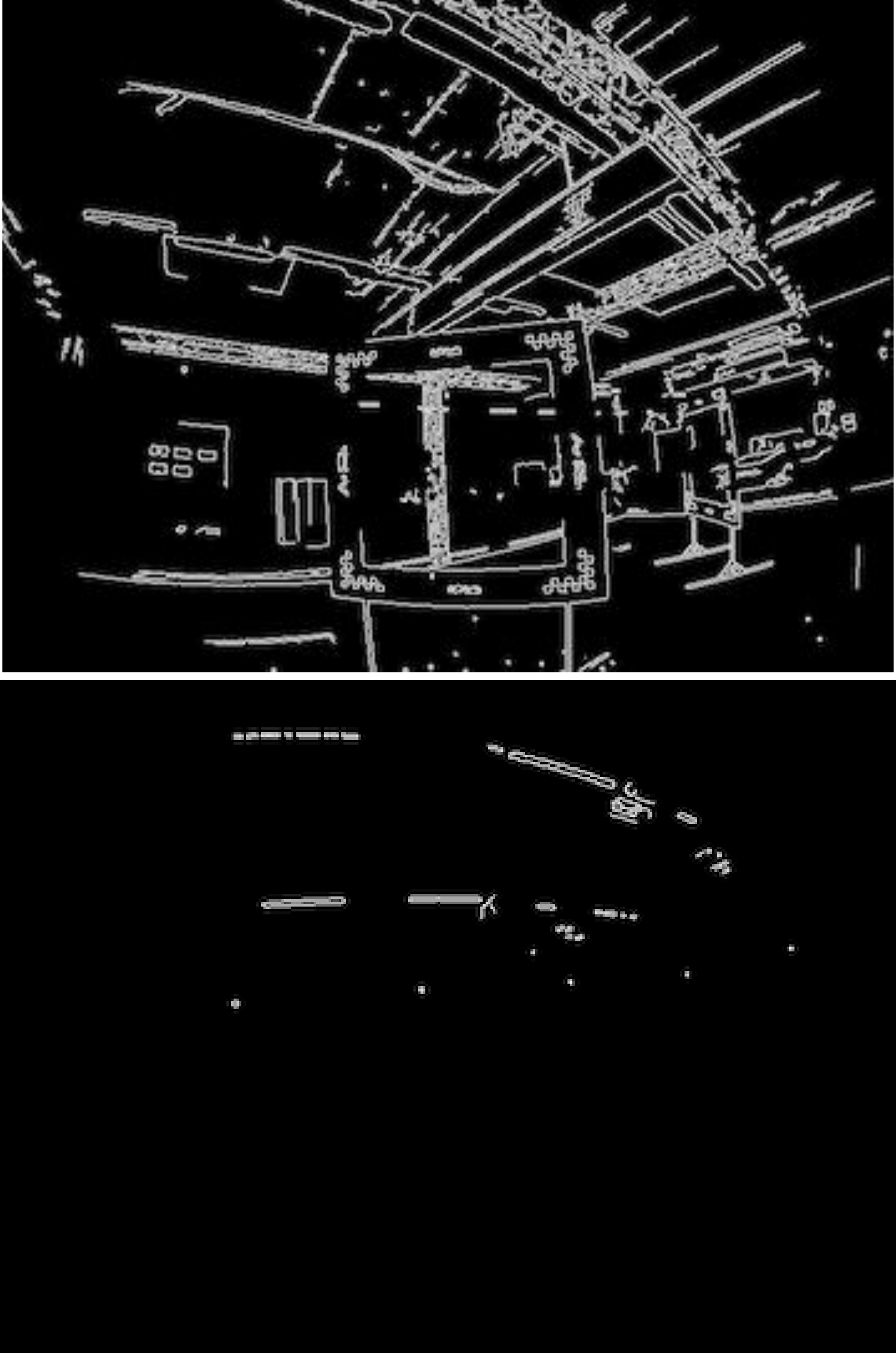}\label{fig:filters_3}}\hfill%
\subfloat[]{\includegraphics[width=0.097\textwidth]{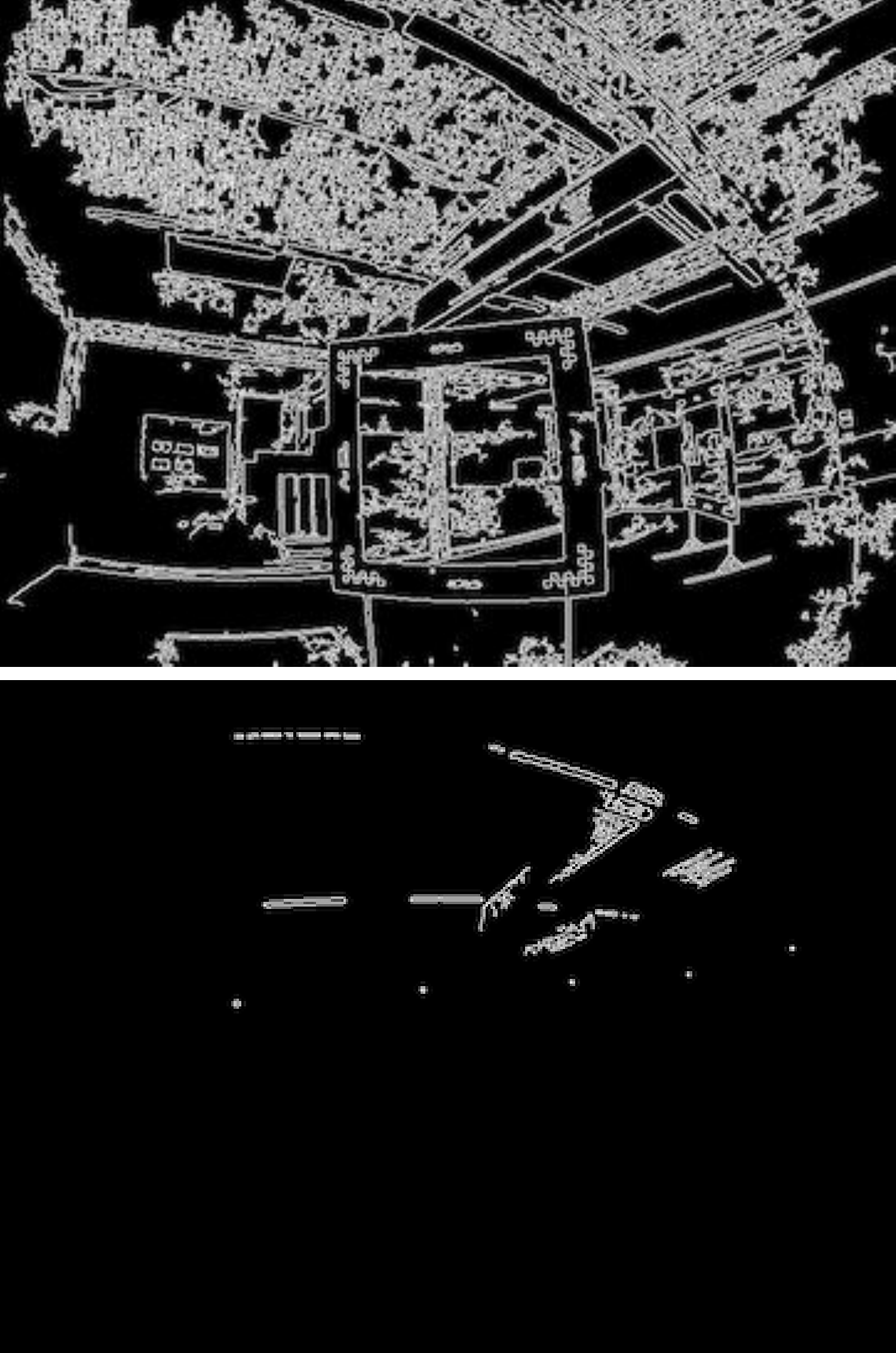}\label{fig:filters_4}}
\caption{Visualization of different filters that are considered and evaluated under bright (the first row) and dark (the second row) lighting: (\ref{fig:filters_0})~original RGB images, and images converted by (\ref{fig:filters_1})~pencil filter, (\ref{fig:filters_2})~Sobel filter, (\ref{fig:filters_3})~Canny filter, and (\ref{fig:filters_4})~Canny filter with a more conservative threshold.}
\label{fig:filters}
\end{figure}

To comprehensively evaluate the effect of the pencil filter, PencilNet is compared against the same backbone network, i.e., GateNet, tailored with other edge detection filters. Three different classical filters are considered, namely the Sobel filter~\cite{Sobel} and Canny filter~\cite{Canny}. Similar to the pencil filter, these filters transform the input images before feeding them to the backbone detection network, as illustrated in Fig.~\ref{fig:filters}.

\subsection{Baselines Training}

As mentioned in Section \ref{sub:sim_data_generation}, PencilNet and the other baseline methods are trained with the same simulation dataset containing $160 \times 120$ RGB images. Note that filter models, such as the proposed method, apply the respective filter on these images before feeding them into its network. All methods use the same training settings, that is set with a batch size of 32, the initial learning rate of $0.01$, and linearly decayed by $0.1$ at epochs 5 and 8. The models are optimized by Adam \cite{kingma2014adam} with default parameters of $\beta_{1}=0.9$, $\beta_{2}=0.999$, and $\epsilon=10^{-8}$.

\subsection{Evaluation Metrics} \label{sub:mectrics}

In the evaluation experiments, mean absolute errors~(MAE) are calculated for the aforementioned methods as a measure of accuracy in predicting gate's center on the image plane ($E_c$), distance to the gate ($E_d$), and orientation of the gate ($E_{\theta}$) relative to the drone's body frame with the ground-truth information as follows:
\begin{equation}
\centering
\begin{cases}
         E_{c}  =  {
         \frac{1}{N} \sum_{i = 1}^{R}
            \sum_{j = 1}^{C}  
    \left( |\hat{x}_{ij} - x_{ij}| + |\hat{y}_{ij} - y_{ij}| \right)}, \\
     E_{d}  = {
            \frac{1}{N} \sum_{i = 1}^{R}
                \sum_{j = 1}^{C}  
       |\hat{d}_{ij} - d_{ij}|}, \\
 E_{\theta} =  {
            \frac{1}{N} \sum_{i = 1}^{R}
                \sum_{j = 1}^{C}  
       |\hat{\theta}_{ij} - \theta_{ij}|},      
    \end{cases}
    \label{eq:metrics}
\end{equation}
where $N$ is the number of test samples.
The metrics in~\eqref{eq:metrics} provide the detection accuracy but do not take into account cases where the model fails to make a prediction and, consequently, produces no errors. Therefore, in addition to the metrics in~\eqref{eq:metrics}, we consider another metric that calculates the percentage of false negative (FN) predictions, i.e., the number of times a network does not detect a gate present in the image over the total number of samples. If a method produces a high FN, it performs poorly even though the other metrics may appear decent.

\subsection{Sim-to-Sim and Sim-to-Real Evaluation}

\remark{By assumption, all gates are in an upright position in all simulation and real-world environments.}

\begin{figure}[!b]
\centering
\subfloat[Good light]{\includegraphics[width=0.12\textwidth]{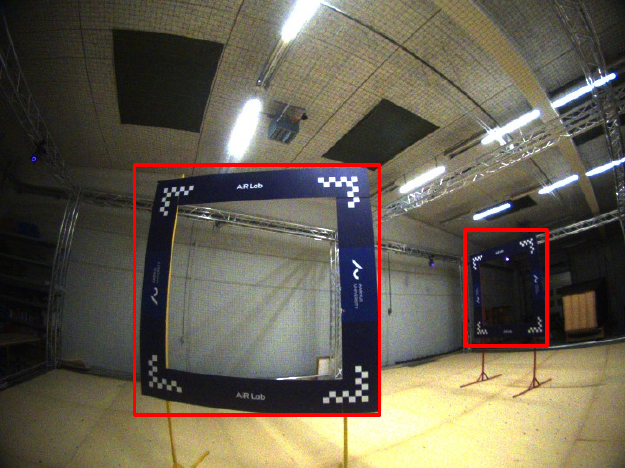}\label{fig:light_high}} \hfill%
\subfloat[Moderate light]{\includegraphics[width=0.12\textwidth]{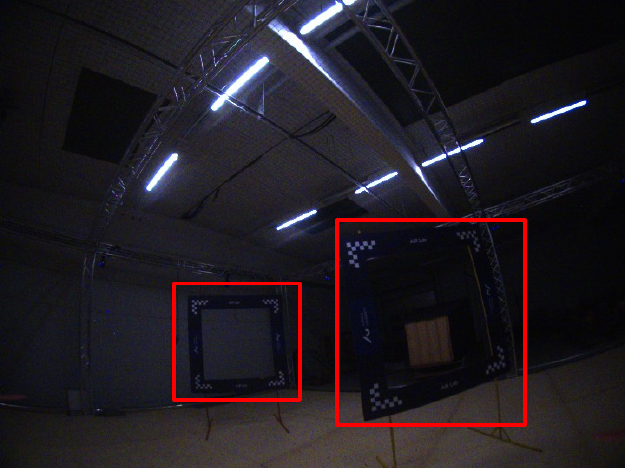}\label{fig:light_medium}} \hfill%
\subfloat[Poor light]{\includegraphics[width=0.12\textwidth]{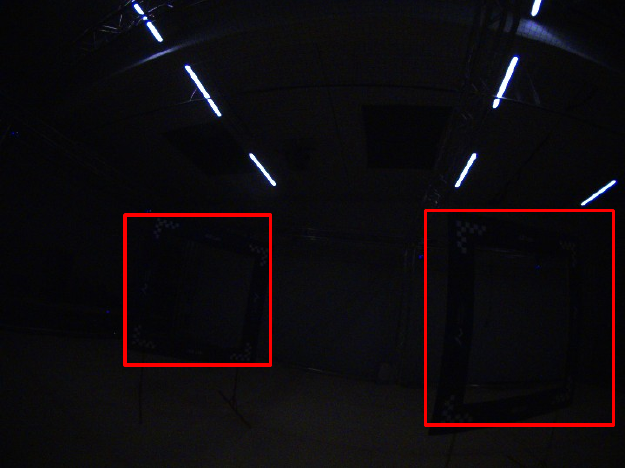}\label{fig:light_low}} \hfill%
\subfloat[Motion blur]{\includegraphics[width=0.12\textwidth]{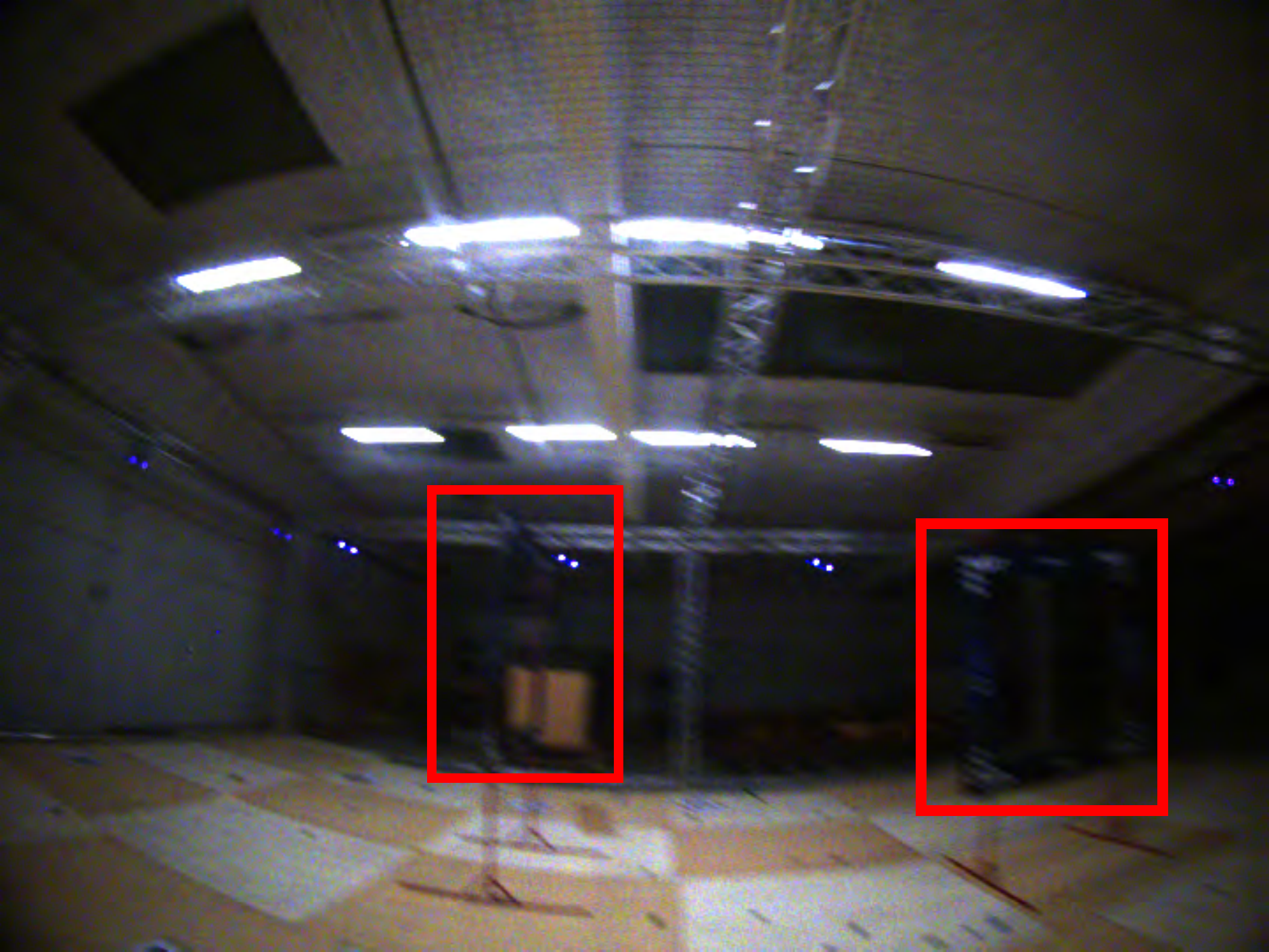}\label{fig:light_blur}}
\caption{Real-time experiment scenes with different conditions such as (a)~high, (b)~medium and (c)~low illumination, and (d)~blurriness.
}
\label{fig:dataset_samples}
\end{figure}

The first objective of the evaluation is to understand how well the models above generalize their understanding of the gate perception task in different environments. For this purpose, all models are evaluated on collected datasets. For the synthetic test dataset (Sim), samples are drawn from three simulation environments that have not been used during the training of PencilNet and other baseline methods. The real-world images, illustrated in Fig.~\ref{fig:dataset_samples}, are obtained from a quadrotor flying fast random trajectories through gates positioned in different layouts at various lighting intensities. The quadrotor is equipped with a high-frequency camera with a fish-eye lens. A motion capture system is used to provide ground-truth information for the poses of the drone and gates. Four real-world datasets (\mbox{\textit{Real N-100}}, \mbox{\textit{Real N-40}}, \mbox{\textit{Real N-20}} and \mbox{\textit{Real N-10}}) are collected at night time with $100\%$, $40\%$, $20\%$ and $10\%$ of artificial light intensity, respectively. Additionally, four datasets (\mbox{\textit{Blur Real N-100}}, \mbox{\textit{Blur Real N-40}}, \mbox{\textit{Blur Real-N20}} and \mbox{\textit{Blur Real N-10}}) are collected containing blur images by increasing both the exposure time of the camera and flight speeds of the drone at different illumination levels. The statistics of the synthetic test dataset are similar to those of the training dataset; while the real-world dataset covers a wide range of situations with great variations in gate poses, as shown in Figs.~\ref{fig:histogram_test}~and~\ref{fig:histogram_real}.

\begin{table*}[t!]
\caption{Comparison of various methods on the test datasets from different environments in terms of the number of parameters (\#p), inference speed in frames per second (fps), accuracy metric ($E_{c}$ in pixels, $E_{d}$ in meters, $E_{\theta}$ in radians) and FN detections in percents.
\label{tbl:comparison}}
\begin{center}
\resizebox{\textwidth}{!}{\begin{tabular}{|c|c|c|cccc|cccc|cccc|cccc|cccc|}\hline
\multirow{2}{*}{Method} & \multirow{2}{*}{\#p} & \multirow{2}{*}{fps} &\multicolumn{4}{c|}{Sim}   &\multicolumn{4}{c|}{Real N-100} &\multicolumn{4}{c|}{Real N-40}  &\multicolumn{4}{c|}{Real N-20}   &\multicolumn{4}{c|}{Real N-10} \\ \cline{4-23}
  
&  & & $E_{c}$ & $E_{d}$ & $E_{\theta}$ & FN & $E_{c}$ & $E_{d}$ & $E_{\theta}$ & FN & $E_{c}$ & $E_{d}$ & $E_{\theta}$ & FN & $E_{c}$ & $E_{d}$ & $E_{\theta}$ & FN & $E_{c}$ & $E_{d}$ & $E_{\theta}$ & FN \\
\hline
\hline

ADRNet-mod \cite{jung2018perception} &2,5M & 14 
& \textbf{0.013} & 0.284 & 0.033 & 24.26
& 0.067 & 1.028 & 0.252 & 19.64
& 0.059 & 0.713 & 0.143 & 86.56
& 0.06 & 0.687 & 0.14 & 98.88
& 0.058 & 0.685 & 0.136 & 97.73 \\

DroNet-1.0 \cite{kaufmann2019beauty} & 478K & 22 
& 0.048 & 0.119 & 0.023 & 62.44
& 0.145 & 0.797 & 0.217 & 61.46
& 0.116 & 0.657 & 0.129 & 99.91
& 0.115 & 0.637 & 0.125 & 99.85
& 0.115 & 0.629 & 0.123 & 99.84\\

DroNet-0.5 \cite{loquercio2019deep} & 250K & 40 
& 0.045 & 0.106 & 0.022 & 51.58
& 0.144 & 0.778 & 0.216 & 47.01
& 0.115 & 0.648 & 0.131 & 98.32
& 0.115 & 0.626 & 0.129 & 99.55
& 0.114 & 0.622 & 0.126 & 99.62\\

GateNet \cite{gatenethuypham2021} & \textbf{32K} & \textbf{57} 
& 0.018 & 0.036 & 0.02 & 24.96 
&  0.071 & 0.323 & 0.228 &15.74
& 0.062 & 0.222 & 0.151 & 66.01 
& 0.066 & 0.234 & 0.145 &65.23 
&  0.064 & 0.238 & 0.139 & 72.07\\

GateNet + sobel & \textbf{32K} & \textbf{57} 
& 0.032 & 0.068 & 0.022 &27.96 
&  0.077 & 0.385 & 0.217 & \textbf{14.32} 
& 0.054 & 0.219 & 0.13 & 79.08 
& 0.054 & 0.219 & \textbf{0.131} & 88.36 
&  0.053 & \textbf{0.223} & \textbf{0.128} & 90.79\\

GateNet + canny & \textbf{32K} & \textbf{57} 
& 0.039 & 0.112 & 0.026 & 78.67
&  0.072 & 0.299 & 0.219 & 95.30
& 0.061 & 0.264 & 0.145 & 90.56
& 0.067 & 0.278 & 0.144 & 93.10
&  0.065 & 0.268 & 0.142 & 89.88\\

%GateNet + laplacian & \textbf{32K} & \textbf{57}
%& 0.056 & 0.29 & 0.036 & 0.24 
%& 0.106 & 0.3 & 0.234 & 0.05
%& 0.077 & 0.321 & 0.158 & 0
%& 0.082 & 0.335 & 0.153 & 0
%&  0.079 & 0.329 & 0.148 & 0\\

\hline \hline
\textbf{PencilNet} & \textbf{32K} & \textbf{57} 
& 0.017 & \textbf{0.031} &  \textbf{0.018} & \textbf{20.5}
&  \textbf{0.065} & \textbf{0.287} & \textbf{0.210} & 18.78
& \textbf{0.051} & \textbf{0.2} & \textbf{0.129} & \textbf{35.10}
& \textbf{0.052} & \textbf{0.209} &  0.135 & \textbf{36}
&  \textbf{0.053} & 0.235 & 0.14 & \textbf{50.18}\\
\hline
\end{tabular}}
\end{center}
\end{table*}
% GateNet 31,580
% ADRNet 2,511,932

\begin{table*}[t!]
\caption{Comparison of various methods on the blur images datasets under different illumination conditions in terms of the number of parameters (\#p), inference speed in frames per second (fps), accuracy metric ($E_{c}$ in pixels, $E_{d}$ in meters, $E_{\theta}$ in radians) and FN detections in percents.
\label{tbl:blur}}
\begin{center}
\resizebox{\textwidth}{!}{
\begin{tabular}{|c|cccc|cccc|cccc|cccc|}\hline
\multirow{2}{*}{Method} &\multicolumn{4}{c|}{Blur Real N-100}  &\multicolumn{4}{c|}{Blur Real N-40}   &\multicolumn{4}{c|}{Blur Real N-20} &\multicolumn{4}{c|}{Blur Real N-10}\\ \cline{2-17}

& $E_{c}$ & $E_{d}$ & $E_{\theta}$ & FN & $E_{c}$ & $E_{d}$ & $E_{\theta}$ & FN & $E_{c}$ & $E_{d}$ & $E_{\theta}$ & FN & $E_{c}$ & $E_{d}$ & $E_{\theta}$ & FN \\
\hline
\hline

%ADRNet-mod \cite{jung2018perception} &2,5M & 14 & 0.051 & 0.806 & 0.134                        & 0.054 & 0.78 & 0.13 & 0.067 & 0.762 & 0.12 & 0.073 & 0.823 & 0.136 \\

%DroNet-1.0 \cite{kaufmann2019beauty} & 478K & 22 & 0.103 & 0.589 & 0.126                                                & 0.123 & 0.681 & 0.121 & 0.127 & 0.723 & 0.121 & 0.136 & 0.629 & 0.772\\

%DroNet-0.5 \cite{loquercio2019deep} & 250K & 40 & 0.102 & 0.583 & 0.123                                                & 0.111 & 0.642 & 0.12 & 0.122 & 0.698 & 0.116 & 0.13 & 0.752 & 0.129\\

GateNet 

& 0.056 & 0.3 & \textbf{0.117} & \textbf{12.45}
&  0.057 & 0.234 & 0.134 & 19.21
& 0.073 & 0.236 & 0.138 & 59.24
& 0.077 & 0.248 & 0.145 & 48.52\\

GateNet + sobel 

& 0.067 & 0.26 & 0.125 & 38.58
& 0.068 & 0.25 & 0.123 & 39.61
& 0.068 & 0.253 & \textbf{0.119} & 82.23
& 0.069 & 0.277 & \textbf{0.129} & 88.41\\

GateNet + canny 

& 0.069 & 0.253 & 0.133 & 85.81
& 0.08 & 0.234 & 0.135 & 80
& 0.099 & 0.275 & 0.135 & 85.99
& 0.101 & 0.281 & 0.147 & 85.17\\

%GateNet + laplacian & 0.082 & 0.369 & 0.154 & 0.08 & 0.356 & 0.141 & 0.08 & 0.339 & 0.144
                                                %& 0.088 & 0.34 & 0.151\\

\hline \hline
\textbf{PencilNet} 
& \textbf{0.048} & \textbf{0.248} & 0.128 & 20.07
& \textbf{0.049} & \textbf{0.225}& \textbf{0.122} & \textbf{14.71}
& \textbf{0.055} & \textbf{0.207} &  0.121 & \textbf{23.69}                                               
& \textbf{0.061} & \textbf{0.215} & 0.137 & \textbf{26.14} \\ 

\hline
\end{tabular}
}
\end{center}
\end{table*}
% GateNet 31,580
% ADRNet 2,511,932

Table~\ref{tbl:comparison} shows the comparison of the gate detection performance in various test environments with different backgrounds and illumination conditions. The number of parameters in each network and their inference speed measured on an NVIDIA Jetson TX2 onboard computer in frames per second~(fps) are compared. The proposed method performs reliably in all environments and illumination conditions. It also outperforms the baseline methods in real-world datasets, particularly in lower illumination. Sobel filter appears to have outstanding performance in the \mbox{\textit{Real-10}} dataset, but it produces a high number of FN predictions and, consequently, performs poorly in practice. Since PencilNet uses GateNet as its backbone network and performs a low-cost operation of pencil filter on small-size images ($\approx 0.3 \si{ms}$ on $160 \times 120$ images), it attains a high inference rate in real-time.

When illumination decreases, all baseline methods predict with lower accuracy and produce a higher number of FN detections. At the same time, the performance of PencilNet retains high accuracy and a low number of FN predictions. PencilNet works well in darker environments is due to the dilation operation in the pencil filter that makes the edges sharper by extracting local maxima within the neighborhood of each pixel. Thus, PencilNet can enhance dim edges and enable the detection of the objects' geometry. Although other edge detection filters used for the GateNet variants could theoretically extract similar features, their performance does not translate well in reality. The main distinction between PencilNet and other GateNet variants is the robustness to noise. By its nature, the Sobel filter is based on  gradient calculation which is more sensitive to the noise in the image. On the other hand, the Canny filter is more robust to noise thanks to the application of non-maximum suppression. However, the hysteresis thresholding requires selecting thresholds that have to be tuned for various light intensities, as illustrated in Figs.~\ref{fig:filters_3}~and~\ref{fig:filters_4}. The local maxima extraction through dilation in the pencil filter does not require parameter tuning. Therefore, PencilNet is more suitable for drone racing applications where sudden illumination changes occur frequently.

On the other hand, PencilNet also performs notably better in sim-to-real schemes compared to other RGB-based models (ADRNet, DronNet variants, and GateNet). Even though being trained on photo-realistic images and domain randomization and having additional color information, it is difficult for these baseline models to understand the real-world environment represented in RGB images. Nevertheless, the loss of color information can negatively affect PencilNet's performance in some situations where the color of the gate can be clearly observable. It is possible to observe from Table~\ref{tbl:comparison}, that PencilNet produces a slightly higher number of FN predictions when evaluated on the real-world dataset \mbox{\textit{Real N-100}} compared to an RGB-based baseline, i.e., GateNet. In this data, the light intensity is maximum. Therefore, the color of the gate is clearly visible, providing an advantage for the RGB-based baseline. However, the rest of the evaluation and experiments illustrate that, in general, learning geometry through enhanced edge detection has better performance.

To further highlight the robustness of the pencil filter over other edge enhancement filter types, Table~\ref{tbl:blur} shows the comparison of different filters for gate detection tasks in real-world blur images that are also commonly observed in drone racing scenarios. PencilNet significantly outperforms other filters in gate center and distance prediction metrics, while having strong performance on orientation prediction with a much lower number of FNs.

\begin{figure}[!b]
\centering
\includegraphics[width=0.49\textwidth]{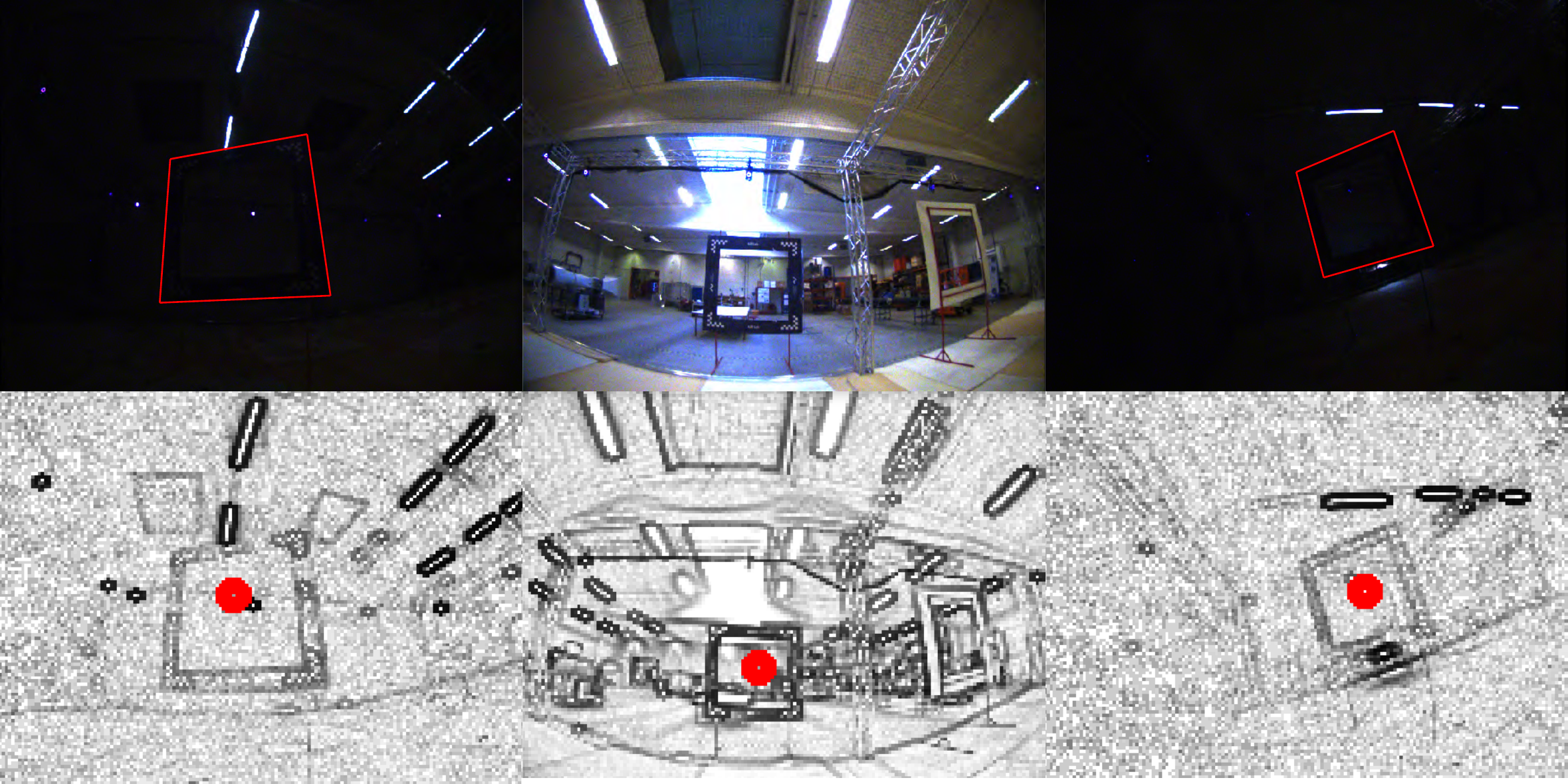}
\caption{Samples of the gate perception by PencilNet. The network can distinguish a gate from objects with similar geometry such as the back of another gate, or in extreme dark lighting.}
\label{fig:sample_prediction}
\end{figure}

A pivotal idea to ensure a successful sim-to-real transfer learning is to narrow the gap between the quality of images rendered by the simulator and the real images obtained from the drone's camera. Thanks to the pencil filter's intelligent abstraction of input data, the network performance is not affected much by ambiance noise and intricate lighting in real-world conditions, as shown in Fig.~\ref{fig:sample_prediction}. This finding is aligned with previous methods that are trained only on simulation data but achieve good real-time performances by adopting a simpler input representation, such as depth map~\cite{loquercio2021learning}. The pencil filter should also be considered as another smart input representation for sim-to-real applications.

\section{Experimental Validation} \label{sec:experiment}

To demonstrate the usefulness of our proposed method, real-world experiments are set up with typical conditions as in recent autonomous drone racing applications, in which a small quadrotor traverses through various sequences of square gates. The drone is equipped with a single RGB camera running at $50 \si{Hz}$ for gate perception, and an Intel Realsense Tracking camera T265 providing pose and velocities estimation. In the following experiments, PencilNet and the baseline methods are run on an NVIDIA Jetson TX2 computer, together with other perception, planning, and control pipelines whose details are provided below.

\subsection{Gate Mapping}

\begin{figure}[!b]
\centering
\includegraphics[width=0.33\textwidth]{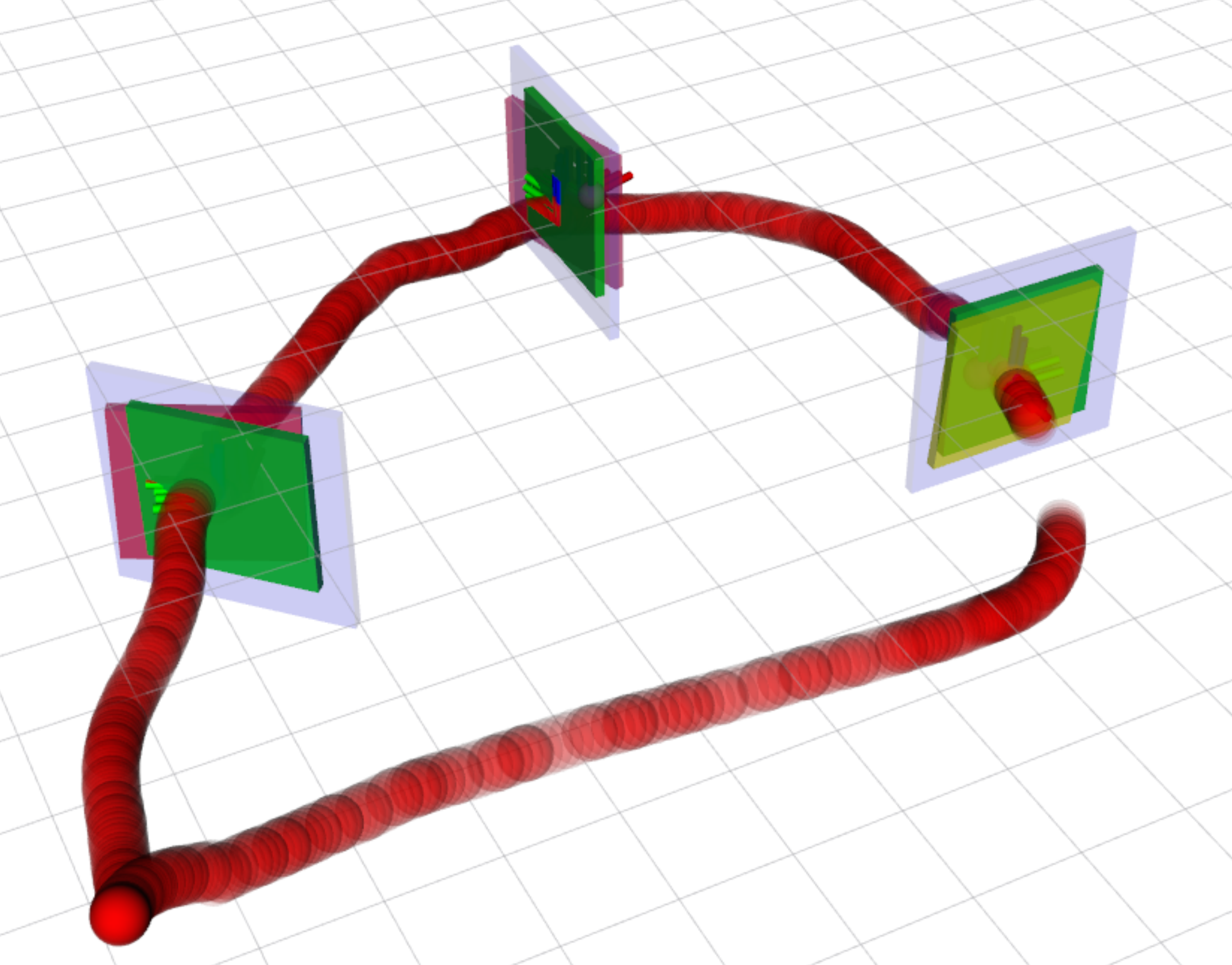}
\caption{Example of the constructed global map. The pink blocks represent the last raw estimation of the gate's pose; while the green blocks represent the estimations by an extended Kalman filter. The generated trajectory is shown with red spheres.}
\label{fig:trajectory_triangle}
\end{figure}

As in similar studies~\cite{kaufmann2019beauty, foehn2020alphapilot, gatenethuypham2021}, we construct a map (as shown in Fig.~\ref{fig:trajectory_triangle}) of the gates in the racing tracks from the output of a perception network, which is a prediction vector $\hat{\mathbf{v}} = \begin{bmatrix} \hat{c}_x & \hat{c}_y & \hat{d} & \hat{\theta} \end{bmatrix}$ of a gate center pixel coordinates $(\hat{c}_x, \hat{c}_y)$, relative distance from the robot to the gate $\hat{d}$, and relative heading angle $\hat{\theta}$. Using the method of fish-eye back-projection proposed in \cite{gatenethuypham2021}, one can obtain the pose of the gate $G$ expressed in world frame $W$ (${}_{W}{\hat{\mathbf{p}}_G}$) as follows:
%%%%%%%%%%%%%%%%%%%%
\begin{equation}
\centering
\begin{cases}
{}_{C}{\hat{\mathbf{p}}_G} = \mathbf{D}^{-1} \mathbf{K}^{-1} \hat{d} \begin{bmatrix} \hat{c}_x, \hat{c}_y, 1 \end{bmatrix}^{T},\\

\begin{split}
&{}_{W}{\hat{\mathbf{p}}_G} = {}_{W}{\mathbf{p}_D} \\
&+ \left( \mathbf{R}^W_D \right)^{-1} \left(  \left( \mathbf{R}^D_C \right)^{-1} \begin{bmatrix} 0 & 0 & 1 \\ -1 & 0 & 0 \\ 0 & -1 & 0 \end{bmatrix} {}_{C}{\hat{\mathbf{p}}_G} + {}_{D}{\mathbf{p}_C} \right).
\end{split}

\end{cases}
\end{equation}
%%%%%%%%%%%%%%%%%%%%
where $W$, $D$, and $C$ are the world, drone’s body, and camera coordinate frames, respectively. $\mathbf{R}^A_B$ expresses the transformation from frame $A$ to frame $B$. $\mathbf{D}$ and $\mathbf{K}$ are the camera distortion matrix and intrinsic matrix. In real-time flights, an extended Kalman filter~(EKF) is used for each detected gate with prior knowledge of coarse gate locations within a radius of six meters on the track, to correctly associate the prediction values with a gate.

\subsection{Motion Planning and Control}

After the global pose of the next gate is estimated, a minimum-snap trajectory is generated~\cite{burri2015real-time} from the drone's current position to a perceived gate's center with a receding-horizon local planner that can replan if the estimation of the next gate pose exceeds a threshold. The trajectory is tracked by a geometric controller\cite{lee2010geometric}.

\subsection{Real-Time Experiments}

\newcommand\colorful{50}

\begin{table}[!b]
\centering
\caption{Success rates of gate crossing with various track layout and illumination settings.}
\resizebox{0.49\textwidth}{!}{
\begin{tabular}{|c|c|c|c|c|c|c|} \hline
\multirow{2}{*}{\textbf{Method}}                                        & \multirow{2}{*}{\textbf{Track}}        & \multicolumn{5}{c|}{\textbf{Lighting}}                                                                                             \\ \hhline{~~-----}
							                                            & 							    & \multicolumn{1}{c|}{\textbf{Day}}  & \multicolumn{1}{c|}{\textbf{N-100}}    & \multicolumn{1}{c|}{\textbf{N-40}}     &    \multicolumn{1}{c|}{\textbf{N-20}} &
							                          \multicolumn{1}{c|}{\textbf{N-10}}       \\ \hline \hline
& {Circular, narrow}     & \cellcolor{green!90!red!\colorful}90\%     & \cellcolor{green!100!red!\colorful}100\%     & \cellcolor{green!90!red!\colorful}90\%   & \cellcolor{green!60!red!\colorful}60\% & \cellcolor{green!10!red!\colorful}10\%  \\ %\hhline{~-----}
& {8-shape, wide} 		& \cellcolor{green!90!red!\colorful}90\%  & \cellcolor{green!100!red!\colorful}100\%  &  \cellcolor{green!65!red!\colorful}75\%  & \cellcolor{green!40!red!\colorful}40\%  & \cellcolor{green!10!red!\colorful}10\% \\ %\hhline{~-----}
\multirow{-3}{*}{{\shortstack{PencilNet\\(ours)}}}               & {S-shape, narrow} 		& \cellcolor{green!95!red!\colorful}95\%  & \cellcolor{green!90!red!\colorful}90\%  & \cellcolor{green!60!red!\colorful}60\%  & \cellcolor{green!10!red!\colorful}10\% & \cellcolor{green!10!red!\colorful}10\%  \\ \hline \hline

& {Circular, narrow}     & \cellcolor{green!60!red!\colorful}60\%  & \cellcolor{green!100!red!\colorful}100\%  & \cellcolor{green!50!red!\colorful}50\%  & \cellcolor{green!30!red!\colorful}30\%  & \cellcolor{green!0!red!\colorful}0\% \\ %\hhline{~-----}
& {8-shape, wide} 		& \cellcolor{green!25!red!\colorful}25\%  & \cellcolor{green!80!red!\colorful}80\% &  \cellcolor{green!13!red!\colorful}12.5\%  & \cellcolor{green!0!red!\colorful}0\%    & \cellcolor{green!0!red!\colorful}0\%  \\ %\hhline{~-----}
\multirow{-3}{*}{{\shortstack{GateNet\\ \cite{gatenethuypham2021}}}}      & {S-shape, narrow} 		& \cellcolor{green!13!red!\colorful}12.5\%  & \cellcolor{green!90!red!\colorful}90\%  & \cellcolor{green!0!red!\colorful}0\%  & \cellcolor{green!0!red!\colorful}0\%    & \cellcolor{green!0!red!\colorful}0\% \\ \hline \hline

& {Circular, narrow}     & \cellcolor{green!60!red!\colorful}60\% & \cellcolor{green!60!red!\colorful}60\%  & \cellcolor{green!0!red!\colorful}0\%  & \cellcolor{green!30!red!\colorful}30\%   & \cellcolor{green!0!red!\colorful}0\%  \\ %\hhline{~-----}
& {8-shape, wide} 		& \cellcolor{green!75!red!\colorful}75\%  & \cellcolor{green!70!red!\colorful}70\%  &  \cellcolor{green!0!red!\colorful}0\%  & \cellcolor{green!0!red!\colorful}0\%   & \cellcolor{green!0!red!\colorful}0\% \\ %\hhline{~-----}
\multirow{-3}{*}{{\shortstack{DroNet-1.0\\ \cite{kaufmann2019beauty}}}}   & {S-shape, narrow} 		& \cellcolor{green!50!red!\colorful}50\%  & \cellcolor{green!60!red!\colorful}60\%  & \cellcolor{green!0!red!\colorful}0\%  & \cellcolor{green!0!red!\colorful}0\%   & \cellcolor{green!0!red!\colorful}0\%  \\ \hline \hline

& {Circular, narrow}     & \cellcolor{green!80!red!\colorful}80\% & \cellcolor{green!60!red!\colorful}60\%  & \cellcolor{green!0!red!\colorful}0\%  & \cellcolor{green!0!red!\colorful}0\%    & \cellcolor{green!0!red!\colorful}0\% \\ %\hhline{~-----}
 & {8-shape, wide} 		& \cellcolor{green!75!red!\colorful}75\%  & \cellcolor{green!80!red!\colorful}80\%  &  \cellcolor{green!13!red!\colorful}12.5\%  & \cellcolor{green!0!red!\colorful}0\%   & \cellcolor{green!0!red!\colorful}0\%  \\ %\hhline{~-----}
\multirow{-3}{*}{{\shortstack{DroNet-0.5\\ \cite{loquercio2019deep}}}}   & {S-shape, narrow} 		& \cellcolor{green!40!red!\colorful}40\%  & \cellcolor{green!60!red!\colorful}60\%  & \cellcolor{green!0!red!\colorful}0\%  & \cellcolor{green!0!red!\colorful}0\%   & \cellcolor{green!0!red!\colorful}0\%  \\ \hline
\end {tabular}}
\label{tbl:experiments}
\end{table}

The vision-based autonomous navigation system is tested in multiple scenarios with different track layouts and backgrounds, as shown in Table~\ref{tbl:experiments}. The level of difficulties for each track is reflected by the time window ($T$), measured by the average gate-to-gate distance and max drone speeds, in which a perception network must predict a gate correctly, and the controller reacts timely. The narrow tracks (Circular and S-shape) have a shorter time window ($T = 1.6 \si{s}$ and $T = 1.16 \si{s}$) so it is more difficult to navigate successfully. The $8$-shape track has a longer time window ($T = 3.25 \si{s}$). The illumination settings of the test environment can be controlled using a led lighting system at night and roof windows during the daytime.

\begin{figure}[!b]
\centering
\includegraphics[width=0.49\textwidth]{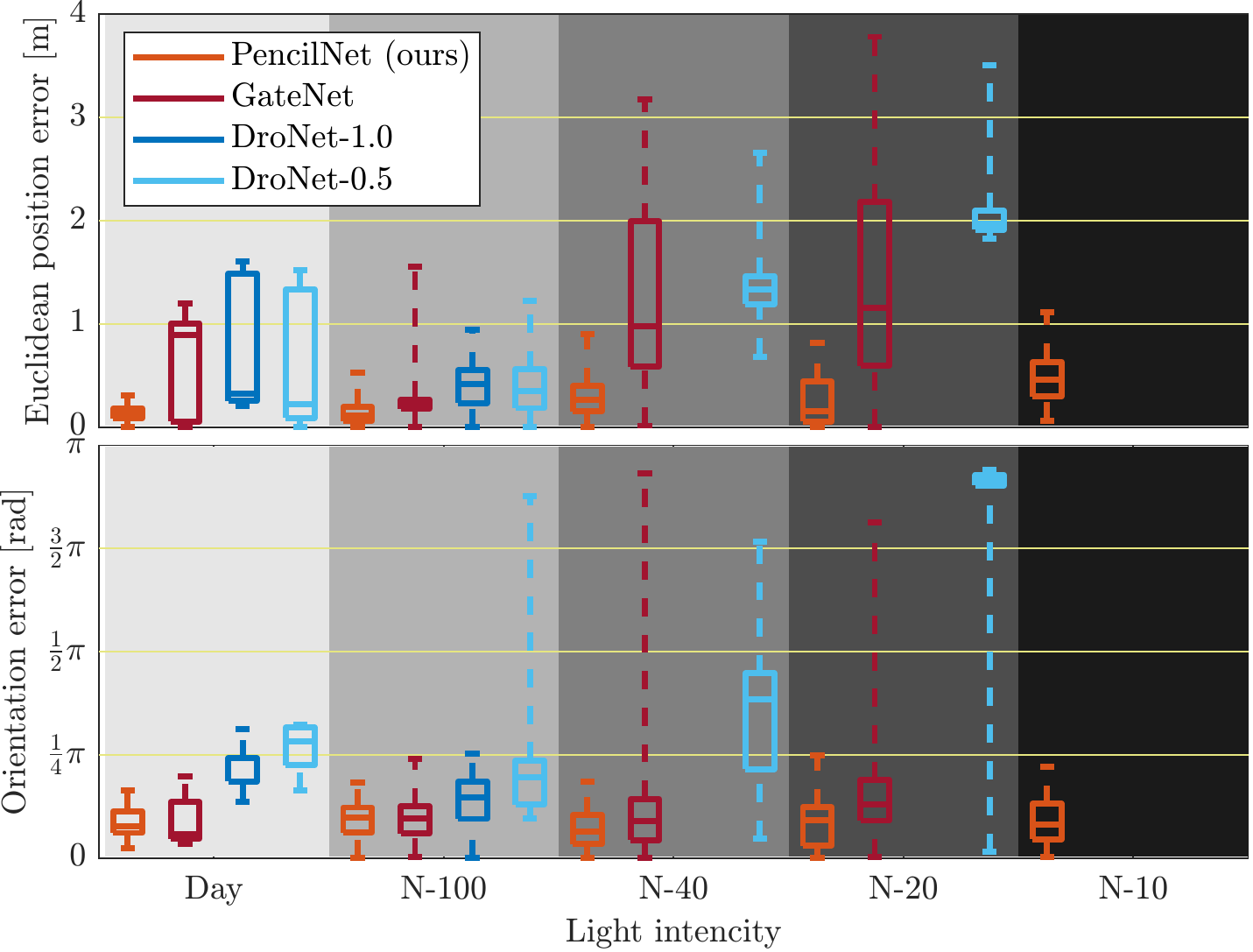}
\caption{Euclidean position and orientation errors for the gate detection with various illumination settings.}
\label{fig:boxplot}
\end{figure} 

The performance of two baseline methods, namely GateNet~\cite{gatenethuypham2021} and DroNet-0.5~\cite{kaufmann2019beauty}, are also reported due to their similarities with our method and their decent speed of inferences. In Table~\ref{tbl:experiments}, we report the success rate for gate crossing for each scenario; while, in Fig.~\ref{fig:boxplot}, the Euclidean errors of each method are depicted. As the inner gate dimension is around 1m, a deviation above $0.5 \si{m}$ normally results in a crash. Since the gates are predicted and then mapped globally, the accuracy of the perception pipeline in the navigation task is directly affected also by the accuracy of the state estimation, which could drift significantly during high-speed flights under degraded visual conditions. This partly explains the poor success rates in the extreme lighting condition (\mbox{N-20}, N-10) for all tested tracks using SLAM. However, the results still show that compared to the similar perception pipeline in~\cite{gatenethuypham2021}, the pencil filter has significantly improved the accuracy of gate detection in practice, especially in darker illumination settings, hence consequently the overall success rate of the mission. Such improvements are valuable in drone racing contests where gates are placed in visually degraded conditions. Readers are encouraged to watch the accompanying video for a detailed review of the performance of the system.

\section{Conclusion}
\label{sec:conclusion}

This work proposes PencilNet -- a DNN method utilizing morphological operation to improve both the sim-to-real transfer learning capability, as well as the robustness of the gate perception system even in extreme illumination settings. Through an extensive evaluation study, PencilNet's performance is shown to significantly outperform state-of-the-art methods for both sim-to-real transfer learning and robustness in detection. We demonstrate the effectiveness of PencilNet in multiple real-time autonomous racing scenarios under different lighting settings that reflect realistic conditions. The performance of the system could be further improved in future by accounting for the drifts and degraded performance in state estimation in darker lighting which negatively affects the accuracy of the perception system.

%\appendix

%\input{sections/projection}

\section*{Acknowledgment}
We would like to thank Micha Heiß, Jonas le Fevre, Abdel Hakim, Jakob Grimm, and Michal Kozlowski for fruitful scientific discussions. This work is supported by Aarhus University, Dept. of Electrical and Computer Eng. (28173) and by the European Union’s Horizon 2020 Research and Innovation Program (OpenDR) under Grant 871449. This publication reflects the authors’ views only. The European Commission is not responsible for any use that may be made of the information it contains.

\bibliographystyle{IEEEtran}
\bibliography{RAL_IROS_2022}

\end{document}